\documentclass{article} 
\usepackage{iclr2026_conference,times}

\usepackage{amsmath,amsfonts,bm}

\def\eqref#1{equation~\ref{#1}}

\def\1{\bm{1}}

\DeclareMathAlphabet{\mathsfit}{\encodingdefault}{\sfdefault}{m}{sl}
\SetMathAlphabet{\mathsfit}{bold}{\encodingdefault}{\sfdefault}{bx}{n}

\usepackage{hyperref}
\usepackage[utf8]{inputenc} 
\usepackage[T1]{fontenc}    
\usepackage{hyperref}       
\usepackage{url}            
\usepackage{booktabs}       
\usepackage{amsfonts}       
\usepackage{nicefrac}       
\usepackage{microtype}      
\usepackage{xcolor}         
\usepackage{amsmath}
\usepackage{svg}
\usepackage{graphicx} 
\graphicspath{ {figure/} }  
\usepackage{algpseudocode}
\usepackage[ruled,vlined,linesnumbered]{algorithm2e}
\usepackage{xcolor} 
\usepackage{multirow}
\usepackage{booktabs}
\usepackage[table]{xcolor}
\usepackage{subcaption}  
\usepackage{caption}
\usepackage[most]{tcolorbox}
\usepackage{adjustbox}
\usepackage{wrapfig}
\usepackage{graphicx}
\usepackage{makecell}
\def\eg{\emph{e.g}.} 
\def\ie{\emph{i.e}.}
\usepackage{listings}
\definecolor{lightgray}{gray}{0.95}
\usepackage[most]{tcolorbox}
\usepackage{tikz}
\newcommand{\cbox}[1]{%
  \tikz[baseline=-0.6ex]{
    \draw (0,0) rectangle (1.2ex,1.2ex);
    \ifnum#1=1
      \draw[line width=1.5pt, line cap=round] (0.2ex,0.6ex) -- (0.5ex,0.3ex) -- (1.0ex,1.1ex);
    \fi
  }%
}
\tcbset{
  myhighlight/.style={
    enhanced,
    breakable,
    colback=yellow!10,
    colframe=yellow!80!black,
    boxrule=0.5pt,
    arc=2pt,
    outer arc=2pt,
    boxsep=5pt,
  }
}
\algrenewcommand\algorithmiccomment[1]{\hfill\textcolor{gray}{$\triangleright$~#1}}

\title{Dynamic Experts Search: Enhancing Reasoning in Mixture-of-Experts LLMs at Test Time}

\iclrfinalcopy

\author{
Yixuan Han$^{1}$\quad
Fan Ma$^{1}$\quad
Ruijie Quan$^{2}$\quad
Yi Yang$^{1}$\footnotemark[2] \quad
\\
$^{1}$ Zhejiang University \quad
$^{2}$ Nanyang Technological University
}

\begin{document}

\maketitle
\begingroup
\renewcommand{\thefootnote}{\fnsymbol{footnote}}
\footnotetext[2]{Corresponding author.}
\endgroup
\begin{abstract}
\label{sec:abstract}
Test-Time Scaling (TTS) enhances the reasoning ability of large language models (LLMs) by allocating additional computation during inference. However, existing approaches primarily rely on output-level sampling while overlooking the role of model architecture. In mainstream Mixture-of-Experts (MoE) LLMs, we observe that varying the number of activated experts yields complementary solution sets with stable accuracy, revealing a new and underexplored source of diversity.
Motivated by this observation, we propose \textbf{D}ynamic \textbf{E}xperts \textbf{S}earch (DES), a TTS strategy that elevates expert activation into a controllable dimension of the search space. DES integrates two key components:
(1) \textit{Dynamic MoE}, which enables direct control of expert counts during inference to generate diverse reasoning trajectories without additional cost; and
(2) \textit{Expert Configuration Inheritance}, which preserves consistent expert counts within a reasoning path while varying them across runs, thereby balancing stability and diversity throughout the search.
Extensive experiments across MoE architectures, verifiers and reasoning benchmarks (\ie, math, code and knowledge) demonstrate that DES reliably outperforms TTS baselines, enhancing accuracy and stability without additional cost.
These results highlight DES as a practical and scalable form of architecture-aware TTS, illustrating how structural flexibility in modern LLMs can advance reasoning.

\end{abstract}

\section{Introduction}
\label{sec:introduction}
Large language models (LLMs) have achieved remarkable progress across
a wide range of
domains~\citep{GPT-4o, o1,Claude,guo2025deepseek,DBLP:journals/corr/abs-2407-21783}, yet their reasoning capability remains a significant challenge~\citep{wei2022chain, Self-Consistency, qi2025mutual, ToT, MindStar}.
Recent advances~\citep{brown2024largelanguagemonkeysscaling,huggingface2024scaling} highlight Test-Time Scaling (TTS) as a promising paradigm to enhance reasoning without additional training: by allocating more computation at inference, TTS generates multiple candidate solutions guided by a verifier and votes one as final solution.
This inference-time paradigm enhances reasoning ability independently of model size, providing an efficient alternative to parameter scaling.

Existing TTS methods typically improve reasoning by expanding the solution space through inference-time search strategies~\citep{wu2024scaling,huggingface2024scaling,qi2025mutual,snell2024scaling, chen2024alphamath,liu2025can}. They rely on stochastic \textbf{sampling} to introduce diversity, increasing the likelihood of finding the correct answer.
However, such diversity is obtained only at the output level, while the internal computation of the model is treated as \textit{architecture-agnostic}. These methods implicitly assume that model architecture plays no role, and therefore treat different architectures in the same way. This assumption becomes problematic in light of recent trends, where many state-of-the-art LLMs adopt the Mixture-of-Experts (MoE) architecture~\citep{SwitchTransformers}. In MoE, a large pool of specialized experts exists, but only a small subset is activated for each input token. Leveraging this flexibility opens up a new dimension for TTS beyond sampling-based diversity alone.

\begin{figure*}[t]
\centering
\includegraphics[width=\textwidth]{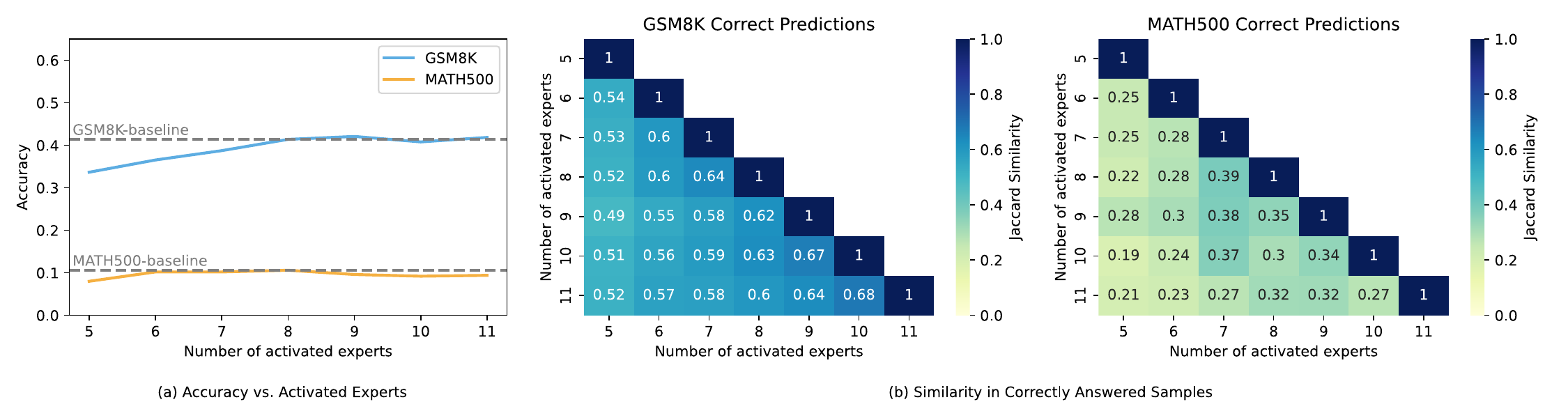}
\caption{
Impact of varying the number of activated experts in MoE models. (a) \textit{Quality}: different activated expert counts yield comparable overall accuracy. (b) \textit{Diversity}: the \textit{Jaccard Similarity} between the sets of problems correctly solved activating different numbers of experts.
}
\label{k-performance}
\vspace{-15pt}
\end{figure*}

From this perspective, we discover an underexplored opportunity for the MoE models: the number of activated experts can be flexibly adjusted at inference, potentially influencing the model's reasoning behavior. To validate this intuition, we study the effect of adjusting expert activation on the \textit{quality} and \textit{diversity} of solutions. As shown in Figure~\ref{k-performance}, varying the number of activated experts yields little change in overall accuracy, but substantially alters the subsets of problems the model solves, with low \textit{Jaccard Similarity} across configurations. In other words, \textit{\textbf{different expert counts lead to complementary solutions}}, highlighting an additional source of diversity beyond output sampling.

Motivated by this observation, we propose \textbf{D}ynamic \textbf{E}xperts \textbf{S}earch (DES), a new TTS strategy that leverages the structural flexibility of MoE architectures now prevalent in state-of-the-art LLMs.
Specifically, DES incorporates two key design choices. The \textbf{first}, \textit{Dynamic MoE}, introduces explicit control over the number of activated experts during inference, treating it as a tunable parameter rather than a fixed default. The \textbf{second}, \textit{Expert Configuration Inheritance}, keeps this number consistent along a reasoning trajectory, allowing the verifier to compare configurations fairly and guide the search toward more effective ones.
Together, these designs elevate expert activation into an additional controllable dimension of the search space: each run maintains a fixed number of activated experts across layers for coherence, while different runs vary this number to explore alternative reasoning trajectories. In this way, DES substantially increases the likelihood of finding correct answers.

To validate this design, we conduct extensive experiments across diverse MoE models and verifiers on a range of reasoning tasks. In our experiments, DES consistently outperforms existing TTS strategies, achieving higher accuracy and precision across math, code, and knowledge benchmarks. Beyond raw performance, DES demonstrates two key advantages: it improves reasoning without increasing computational cost, and it generalizes effectively across different model scales and verifier choices. These results highlight that treating expert activation as a controllable dimension provides a practical and scalable pathway to stronger reasoning in MoE-based LLMs. More broadly, DES establishes a paradigm of architecture-aware TTS and illustrates how structural flexibility in modern LLMs can be systematically harnessed to advance reasoning capabilities.

Our contributions can be summarized as follows:
\begin{itemize}

\item We identify the limitation of existing TTS strategies that rely solely on output-level sampling and reveal expert activation in MoE models as a new axis of exploration for reasoning.

\item We propose DES, a new TTS strategy designed for MoE models that treats expert activation as a controllable dimension of the search space. DES integrates two design choices to explore different expert activation, increasing the probability of obtaining the correct answer.

\item Extensive experiments across diverse MoE models, verifiers, and reasoning benchmarks show that DES outperforms existing TTS methods at comparable cost, highlighting architectural diversity as a practical and scalable pathway to stronger reasoning in LLMs.

\end{itemize}

\section{Related Work}
\label{sec:related_work}

\textbf{Test-Time Scaling.}
Scaling the test-time computation of LLMs has been shown to be an effective method to enhance model performance~\citep{brown2024largelanguagemonkeysscaling}. In prior work, Deepseek-R1~\citep{guo2025deepseek} promoted the generation of long chain-of-thought (Long CoT) during training, thereby guiding the model to increase its computational efforts at inference time, which effectively improves reasoning capabilities. A variety of techniques operating exclusively during inference have also been explored, including majority voting~\citep{Self-Consistency}, search-based strategies~\citep{wu2024scaling,ToT, xie2024self, wan2024alphazero, qi2025mutual}, and iterative refinement~\citep{RISE}. These approaches involve generating multiple candidate responses for a given question and subsequently selecting the final answer through voting or optimization. Subsequent works~\citep{MindStar, wu2024scaling, snell2024scaling} have extended this method by incorporating process-level rewards to guide the search.~\cite{huggingface2024scaling} proposed enhancing TTS through search methods that explicitly promote diversity among sampled outputs. Furthermore, ~\cite{liu2025can} performed a comprehensive evaluation of various TTS methods, demonstrating that the optimal scaling strategy is based on the policy models, PRMs, and difficulty levels of the problem.~\cite{xu2025phidecodingadaptiveforesightsampling} proposed a search method based on foresight sampling, improving performance through token-level reward feedback foresight. However, a common limitation of these works is their reliance on temperature sampling as the sole source of output diversity. Our method leverages the scalability of MoE architectures to achieve a broader exploration, effectively improving reasoning performance.

\textbf{Mixture-of-Experts.}
MoE~\citep{shazeer2017} paradigm has gained significant attention in recent years due to its ability to improve model specialization and computational efficiency~\citep{jin2025moe,wang2025remoe}. MoE architectures, which were originally proposed by~\cite{Jacobs_Jordan_Nowlan_Hinton_1991, shazeer2017}, have been widely explored to tackle complex tasks utilizing multiple expert models in a modular and collaborative manner~\citep{huang2025jakiroboostingspeculativedecoding}. MoE architectures have shown substantial promise in scaling LLMs, where computational efficiency is paramount. Recent advances in MoE have been made in a variety of domains. GShard~\citep{lepikhin2021gshard} and Switch Transformer~\citep{SwitchTransformers} demonstrated the scalability of MoE models by using top-k routing strategies, allowing large models to be trained with a sparse activation of experts. In terms of optimizing MoE architectures, several studies~\citep{DBLP:journals/corr/abs-2403-07652, li2023adaptive, anonymous2024adamoe, jin2025moe} have explored the dynamic selection of experts to enhance diversity and specialization, and some works explored how to adequately use the capacity of MoE models~\citep{lewis2021baselayerssimplifyingtraining,Roller_Sukhbaatar_Szlam_Weston_2021,zhou2022mixtureofexperts}. Beyond optimization, MoE's inherent flexibility also has been utilized to decouple token-level predictions and improve diversity by~\cite{huang2025jakiroboostingspeculativedecoding} in speculative decoding. By incorporating MoE architecture into an advanced search strategy, we explore new avenues to enhance its effectiveness in test-time scaling, with a focus on expanding the exploration space, refining the search process, and strengthening the reasoning capacity of models.

\section{Method}
\label{sec:method}

\subsection{Problem Definition}
\label{preliminary}

\textbf{Test-Time Scaling.}
Given a question $q$, TTS strategy utilizes a policy model $\pi_{\theta}$ and a verifier $R_{\phi}$ to solve it under a given computational budget $N$(\eg, generate $N$ candidate solutions) with a maximum number of reasoning steps $T$. Let \(\mathcal{S} = \{s_0, s_1, \cdots , s_T\} \) be the state set with $s_t$ denoting the result after the $t$ steps reasoning. Similarly, let \(\mathcal{A}=\{A_0, \cdots, A_T \}\), where \({A}_t=\{a_{t}^{(0)},\cdots, a_{t}^{(m-1)}\}\) denotes the candidate action set at the $t$-th step and $a_{t}^{(m)}$ denoting the $m$-th candidate intermediate step generated at $t$-th step. The process begins with the initial state \(s_0 = q\). Policy model \(\pi_\theta\) iteratively generates $m$ actions \(a_{t}^{(i)}\sim\pi_\theta(\cdot \mid s_t)\), \(i \in [0,m)\) at each timestep \(t\). Then the verifier produces a scalar reward \(r_{t}^{(m)} = R_{\phi}(s_t, a_{t}^{(m)})\) for each action, and the state evolves with the best action through deterministic concatenation: \(s_{t+1} = [s_t, a_{t}^{(argmax(r^{(i)}_{t}))}]\). This iterative interaction ends when reach maximum steps \(T\) or an explicit termination signal (\texttt{<EOS>}) is produced. 

\textbf{Vanilla MoE Inference.} 
MoE architectures implement sparse computation by selectively activating a subset of expert networks at each layer~\citep{shazeer2017}.
A typical MoE layer consists of a router and a set of lightweight feed-forward networks (experts)~\citep{SwitchTransformers}. 
During inference, the router selects a fixed number of top-scoring experts for each input token based on its representation.
This enables token-level specialization while maintaining computational efficiency. 
Building on this, we identify two often overlooked properties of MoE inference: (1) small variations in the number of activated experts have negligible impact on overall accuracy, and (2) different activation counts can solve distinct, partially non-overlapping sets of problems.

\begin{figure*}[t]
    \centering
    \includegraphics[width=\textwidth]{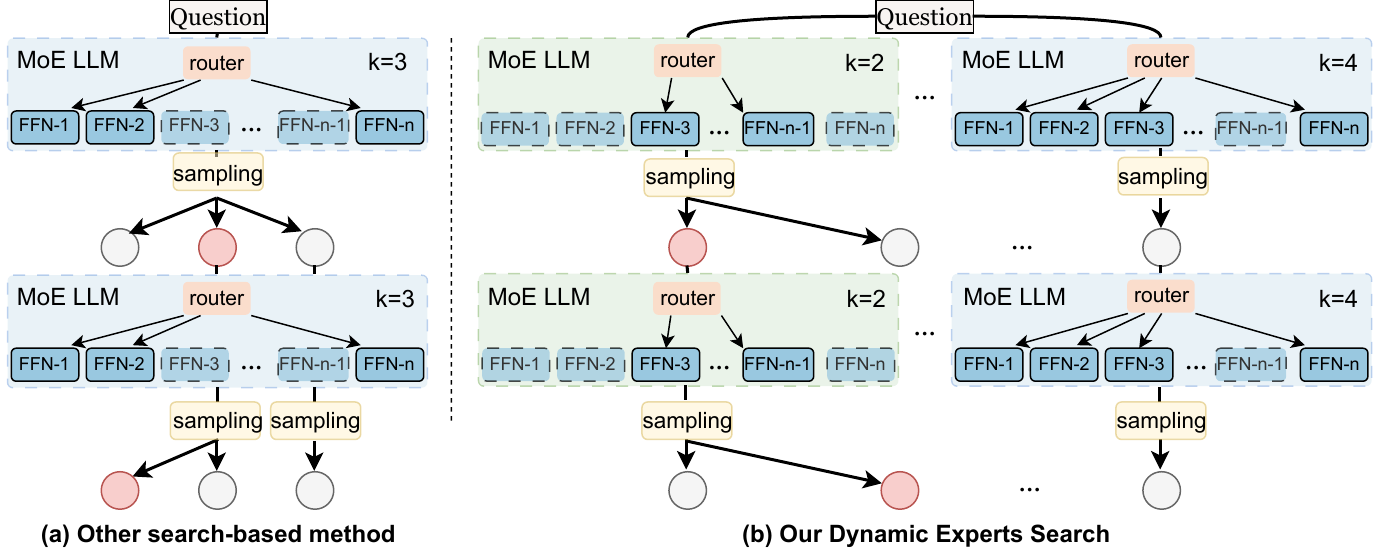}
    \caption{Comparison of other search-based method and our Dynamic Experts Search when applied on MoE models. \includegraphics[height=1em]{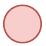} denotes the correct step and \includegraphics[height=1em]{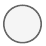} represents an incorrect one. }
    \label{moe-comparison}
\vspace{-15pt}
\end{figure*}

\subsection{Dynamic Experts Search}
In this section, we propose \textbf{D}ynamic \textbf{E}xperts \textbf{S}earch (DES), 
which dynamically explores different expert activation counts within MoE models. As illustrated in Figure~\ref{moe-comparison}, TTS search strategies solve a question step-by-step(\ie, generating an intermediate step at each timestep $t$). However, when applied to MoE models, existing approaches activate only a fixed number of experts throughout the process, whereas our DES explores varying expert counts. Our DES consists of \textit{Dynamic MoE} and \textit{Expert Configuration Inheritance}. Together, these two components form a unified strategy for dynamic expert selection during multi-step reasoning. Below, we describe each module in detail.

\textbf{Dynamic MoE.}
To introduce expert variability into the TTS search process, we augment the MoE model with the ability to control the number of activated experts.
In standard MoE settings, this number is typically fixed and hard-coded across all layers.
In contrast, our \textit{Dynamic MoE} mechanism treats the activation count \(k\) as a tunable parameter during inference.
We define the policy model $\pi_{\theta}$ to take as input a state $s$ and an expert activation counts $k$, and generate the next intermediate reasoning step conditioned on state $s$ by activating $k$ experts for each forward pass.
Formally, the intermediate steps are sampled from $a \sim \pi_{\theta} ( \cdot \mid s, k)$. 

\textbf{Expert Configuration Inheritance.}
When solving a question step by step, DES leverages Dynamic MoE to explore multiple expert activation counts.
As evidenced by the observations in Figure~\ref{k-performance}, each question tends to have an optimal expert activation count. Consequently, uniformly exploring all activation counts throughout the entire search process inevitably wastes part of the computation budget on suboptimal choices.
To address this, we introduce \textit{Expert Configuration Inheritance}, which enables the search to progressively focus on the most promising activation count. 
Specifically, it ensures that for a given state $s_t = [s_{t-1}, a_{t-1}]$, the activation count used to expand $s_t$ is the same as that used to expand $s_{t-1}$ (\ie, to generate $a_{t-1}$). 
This inheritance mechanism ensures that the number of experts remains consistent along each reasoning trajectory.
At the first step, the model uniformly explores a predefined set of expert activation counts. As the search progresses, the verifier assigns lower scores to less promising activation counts, making them unlikely to be inherited in subsequent steps. In this way, more computational resources are allocated to activation counts that lead to higher rewards, enabling focused exploration around promising activation counts and increasing the likelihood of discovering correct answers.

\textbf{Dynamic Experts Search.}
Given a question $q$ with a computation budget $N$ (\ie, generate $N$ rollouts per problem), at timestep $t = 0$ we initialize the search with a predefined set of expert activation counts \(\{k_0, \cdots, k_{n-1}\}\) with $k_i\neq k_j$ when $i \neq j$ and construct the initial candidate set
\begin{equation}
  C_0 = \{(s_0, k_0), \cdots, (s_0, k_{n-1})\}.  
\end{equation}
Here, $(s_0, k_i)$ denotes that given state $s_0 = q$, the next reasoning step will be generated by activating $k_i$ experts for each forward pass of \textit{Dynamic MoE} model $\pi_{\theta}$. To satisfy the computation budget $N$, each activation count $k_i$ is allocated to generate \(|\frac{N}{n}|\) intermediate steps. Consequently, we can obtain a set of $N$ candidate intermediate steps, each paired with its corresponding expert activation count:
\begin{equation}
\label{eq2}
 A_0 =  \{(a^{(j)}_0, k_i) \mid \, i \in [0,n), \, j \in [0, \frac{N}{n}), \, a^{(j)}_0 \sim \pi_{\theta}(\cdot \mid q, k_i)\},
\end{equation}
where $a^{(j)}_0 \sim \pi_{\theta}(\cdot \mid q, k_i)$ denotes the $j$-th single sample drawn from $\pi_{\theta}(\cdot \mid q, k_i)$ when $t=0$.
Each intermediate step in $A_0$ can serve as the first step to solve $q$. We concatenate $s_0$ with each candidate intermediate step to form $N$ candidate states paired with corresponding expert activation count:
\begin{equation}
    \label{eq3}
    C_0' = \{ ([s_0, a], k) \mid \, (a,k) \in A_0\}.
\end{equation}
Then we score every states in $C_0'$ using the verifier $R_{\phi}$ and retain the $\text{top}-M$ states according to these scores where $M$ is a predefined parameter:
\begin{equation}
    \label{eq4}
    C_1 = \{(s, k) \in C_0' \mid R_{\phi}(s) \in Top_M(R_{\phi}(C_0'))\} = \{(s_t^{(0)}, k_0), \cdots, (s_t^{(M-1)},k_{M-1})\}.
\end{equation}
In this process, the \textit{Expert Configuration Inheritance} mechanism records the number of activated experts used to generate each intermediate step and propagates this information into $C_1$, ensuring that each candidate preserves its activation count when generating the subsequent step. The set $C_1$ then serves as the starting state for the next reasoning step. Similarly, at timestep $t$, the set $C_t$ received from timestep $t-1$ undergoes the operations described in Equations~\ref{eq2}~\ref{eq3}~\ref{eq4} to generate $C_{t+1}$, and this procedure continues until either the maximum reasoning step is reached or the end-of-sequence token ("\texttt{<EOS>}") is generated. The detailed process of DES is shown in the Algorithm~\ref{algorithm}. Upon completion of the search, we obtain $N$ complete answers to the question $q$. Then a single answer is selected—the one achieving the highest verifier score or the highest majority vote—as the final answer.
 
\begin{algorithm}[t]
\small
\caption{Dynamic Experts Search}
\label{algorithm}
\SetKwInOut{Input}{Input}
\SetKwInOut{Output}{Output}

\Input{Question $q$, computational budget $N$, candidates counts retained per step $M$, dynamic MoE policy model $\pi_\theta$, verifier $R_\phi$, maximum steps $T$, different numbers of activated experts $\{k_0,\cdots,k_{n-1}\}$ with $k_i\neq k_j$ when $i \neq j$ } 

Initialize $s_0 \leftarrow q$, timestep $t \leftarrow 0$, priority queue $C_0 \leftarrow [\,\,] $;

\For{$i=0$ \textnormal{to} $n-1$}{
Add $(s_0, k_i)$ to $C_0$ \Comment{initialize candidates}
}

\While{$t < T$ \textnormal{and} \textnormal{no end-of-sequence token in} $C_t$}{
  
  \For{ $(s,k) \in C_t$}{
    Sample actions $\{a^{(j)}\}_{j=0}^{ \frac{N}{|C_t|}-1} \sim \pi_\theta(\cdot|s,k)$;\Comment{Sample using $k$ experts}
    
    \For{$j = 0$ \textnormal{to} $\frac{N}{|C_t|}-1$}{
      Add $([s, a^{(j)}], k)$ into $C_{t}'$\; 
    }
  }
  $C_{t+1} \leftarrow \{(s, k) \in C_t' \mid R_{\phi}(s) \in Top_M(R_{\phi}(C_t'))\}$; \Comment{Update candidates}
  
  $t \leftarrow t + 1$\;
}

$(\hat{s}, \hat{k}) \leftarrow \arg\max\limits_{(s,k) \in C_T} R_{\phi}(s)$\;

\Return{$\hat{s}$}; \Comment{Return solution with highest reward}
\end{algorithm}

\textbf{Discussion.}
DES introduces a structured and adaptive approach to inference-time exploration that differs fundamentally from existing TTS strategies (as shown in Figure~\ref{moe-comparison}). We highlight the following key distinctions:
(1) Traditional TTS methods overlook that expert selection within MoE models could be exploited to broaden the capability boundaries of MoE models. In contrast, DES leverages \textit{\textbf{Dynamic MoE}} to vary the number of activated experts, which introduces structural diversity in the computational pathway of models and enlarges the exploration space for identifying correct answers.
(2) Through \textit{\textbf{Expert Configuration Inheritance}}, DES maintains consistent expert activation counts across each reasoning path. This design enables implicit configuration-level credit assignment and allows the model to gradually converge to the most effective expert activation count for the task.

\section{Experiments}
\label{sec:experiments}
\begin{table*}[t]
\centering
\caption{\texttt{Accuracy} (Acc $\uparrow$) and \texttt{Precision} (Prec $\uparrow$) of different strategies on benchmarks when using \texttt{Qwen2.5-Math-PRM-7B} as policy model. For implementation, we generate $N=32$ rollouts for each problem. Additional results for other models are provided in the \textbf{Appendix~\ref{appendix:a}}.}
\label{tab:main_table}
    \resizebox{1.0\columnwidth}{!}{
        \setlength\tabcolsep{6pt}
        \renewcommand\arraystretch{1.0}
\begin{tabular}{lccccccc}
\toprule

\textbf{Strategy} & \textbf{Metric} & \textbf{MATH500} & \textbf{AIME24} & \textbf{AIME25} & \textbf{HumanEval} & \makecell{\textbf{LiveCodeBench}\\(v6-lite)}  & \makecell{\textbf{LiveBench}\\(reasoning)} \\
\midrule
\rowcolor{gray!20}
\multicolumn{8}{c}{\texttt{\textbf{Qwen3-30B-A3B-Instruct}}} \\
\midrule
\multirow{2}{*}{Best-of-N}  
        & Acc(\%)    & 92.40 & 83.33& 66.67  & 92.07 & 35.11 & 90.50 \\
        & Prec(\%)   & 91.38 & \textbf{73.02} & 56.25  & 92.23 & 31.99 & 79.34\\
\midrule
\multirow{2}{*}{BeamSearch} 
        & Acc(\%)    & 93.00 & 83.33 & 63.33 & 89.02 & \textbf{37.40} & 91.50 \\
        & Prec(\%)   & 92.16 & 72.08 & 54.79 & 93.48 & 31.82 & 79.11 \\
\midrule
\multirow{2}{*}{DVTS}      
        & Acc(\%)    & 87.40 & 86.67 & 70.00  & 93.90 & 32.06 & 90.00 \\
        & Prec(\%)   & 83.60 & 71.98 & 57.92 & 92.85 & 32.03 & 79.17 \\
\midrule
\multirow{2}{*}{\textbf{DES(Ours)}} 
        & Acc(\%)    & \textbf{93.20} & \textbf{86.67} & \textbf{70.00}  & \textbf{94.51} & 33.59 & \textbf{91.50} \\
        & Prec(\%)   & \textbf{92.20} & 72.29 & \textbf{58.44} & \textbf{93.75} & \textbf{32.94} &  \textbf{81.58} \\
\midrule
\rowcolor{gray!20}
\multicolumn{8}{c}{\texttt{\textbf{Ling-lite-1.5}}} \\
\midrule
\multirow{2}{*}{Best-of-N}  
        & Acc(\%)    & 80.00 & 26.67 & 23.33 & 79.88 & 19.08 & 22.00 \\
        & Prec(\%)   & 77.72 & 17.19 & 13.65 & 82.13 & 17.65 & 13.58 \\
\midrule
\multirow{2}{*}{BeamSearch} 
        & Acc(\%)    & 81.60 & 23.33 & 13.33 & 82.32 & 19.84 & 20.50 \\
        & Prec(\%)   & \textbf{78.46} & 16.56 & 11.77 & 83.38 & \textbf{19.13} & 12.92 \\
\midrule
\multirow{2}{*}{DVTS}    
        & Acc(\%)    & 82.80 & 26.67 & \textbf{23.33} & 82.93 & \textbf{20.61} & 21.00 \\
        & Prec(\%)   & 77.19 & \textbf{20.00} & \textbf{15.94} & \textbf{84.60} & 18.89 & 13.73 \\
\midrule
\multirow{2}{*}{\textbf{DES(Ours)}} 
        & Acc(\%)    & \textbf{83.80} & \textbf{26.67} & 16.33 & \textbf{83.54} & 19.84 & \textbf{23.00} \\
        & Prec(\%)   & 77.27 & 19.27 & 12.29 & 84.18 & 17.65 & \textbf{14.39} \\
\bottomrule
\end{tabular}
}
\vspace{-5pt}
\end{table*}
\begin{figure*}[t]
\centering
\includegraphics[width=0.9\textwidth]{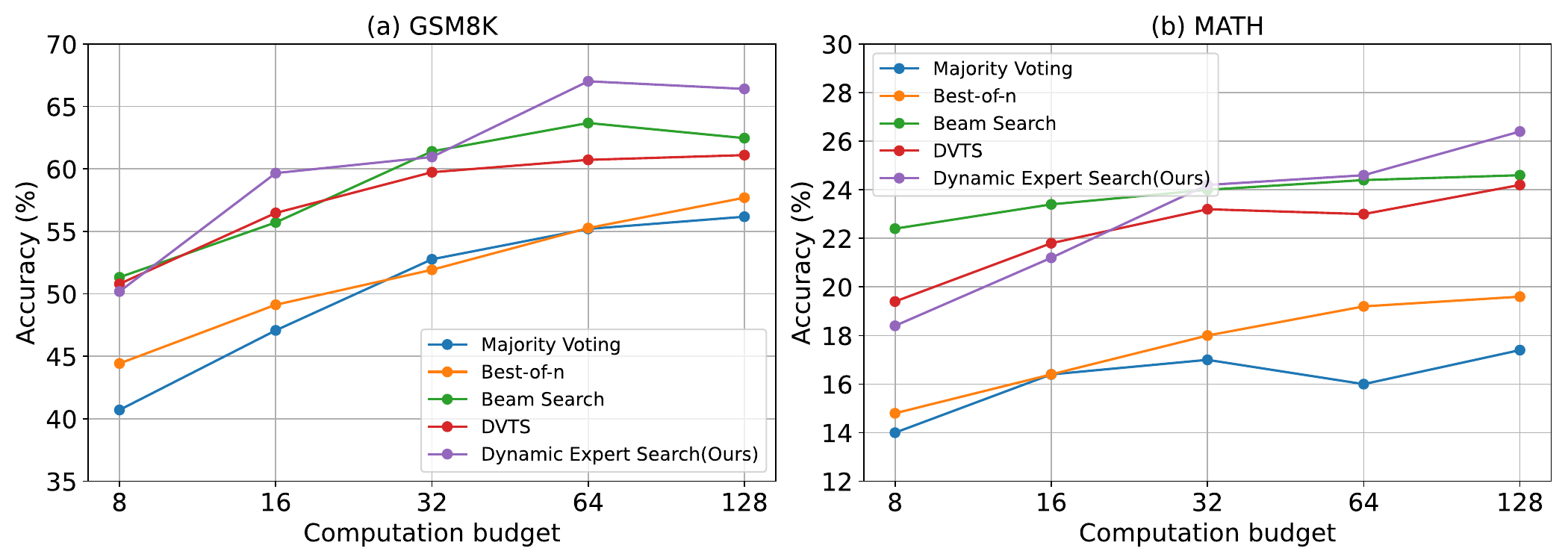}
\caption{Comparison of different search strategies applied to \texttt{OLMoE-1B-7B-Instruct \& Llama3.1-8B-PRM-Deepseek-Data}.
(a) Results on GSM8K. (b) Results on MATH-500.}
\label{search_strategy_comparison}
\vspace{-15pt}
\end{figure*}

\subsection{Setup}
\label{subsec:setup}
\textbf{Models and Benchmarks.}
We select four MoE-based policy models: \textbf{\texttt{Qwen3-30B-A3B}}~\citep{qwen3technicalreport}, \textbf{\texttt{Ling-lite-1.5}}~\citep{ling}, \textbf{\texttt{OLMoE-1B-7B-Instruct}}~\citep{muennighoff2025olmoe} and \textbf{\texttt{DeepSeek-V2-Lite-Chat}}~\citep{deepseekv2}. We utilize two widely used process reward models which are designed capable of scoring each intermediate reasoning step within a multi-step solution as verifier: \textbf{\texttt{Qwen2.5-Math-PRM-7B}}~\citep{anonymous2025the} and \textbf{\texttt{Llama3.1-8B-PRM-Deepseek-Data}}~\citep{wang2024math}. In this work, we conduct evaluations on a several widely used reasoning benchmarks in the \textit{math}, \textit{code} and \textit{knowledge} domains to examine the reasoning performance of DES in the \textbf{0-shot} setting. For the math domain, we evaluate on \textbf{\texttt{MATH500}}~\citep{PRM800K}, \textbf{\texttt{AIME24}}~\citep{AIME24} and \textbf{\texttt{AIME25}}~\citep{AIME25} (\texttt{\textbf{GSM8K}}~\citep{GSM8K}, \texttt{\textbf{SVAMP}}~\citep{saha-etal-2021-svamp} are used for models under 16B), all of which consist of diverse mathematical reasoning problems written in natural language. For the code domain, we evaluate on \textbf{\texttt{HumanEval}}~\citep{chen2021evaluating} and \textbf{\texttt{LiveCodeBench(v6-lite)}}~\citep{jain2025livecodebench}, both of which cover a wide range of programming tasks. For the knowledge domain, we evaluate on \textbf{\texttt{LiveBench(reason)}}~\citep{white2025livebench}, which require knowledge-based reasoning.

\textbf{Baselines and Metrics.} We compare DES with three baseline search strategies. \texttt{\textbf{Best-of-N}}~\citep{brown2024largelanguagemonkeysscaling} samples $N$ complete responses and selects the solution with highest reward scored by process reward model as the final answer. \texttt{\textbf{Beam Search}}~\citep{snell2024scaling} generates $N$ candidates per step, retains the top-$M$ scored by verifier, and expands each with $\frac{N}{M}$ new steps iteratively. After reaching the maximum reasoning step or the termination signal being produced, Beam Search selects final answer via scoring by verifier. \texttt{\textbf{Diverse Verifier Tree Search}}~\citep{huggingface2024scaling} enhances diversity by partitioning the search into $M$ independent subtrees, each searched separately using Beam Search. We report \texttt{\textbf{Accuracy}} (Acc $\uparrow$) for different search strategies, which refers to the proportion of correctly solved problems when the final answer is determined by \textbf{majority voting} (\ie, selecting the most frequently occurring response). We also report \texttt{\textbf{Precision}} (Prec $\uparrow$) for each search strategy on each benchmark, which indicates the fraction of correct answers among the $N$ candidate solutions returned by the search process. In addition, we include the \texttt{\textbf{Average number of generated tokens per problem}} (\#Gen.Tok $\downarrow$) as a measure of the computational cost.

\begin{table*}[t]
\centering
\caption{\texttt{Accuracy} (Acc $\uparrow$) of different modes for \texttt{Qwen3-30B-A3B}. We evaluate four modes: with think-mode, with DES, with instruction tuning and with DES + instruction tuning. For DES, we generated $N=32$ rollouts for each problem and use \texttt{Qwen2.5-Math-PRM-7B} as verifier.}
\label{tab:compare_with_thinkmode}
\resizebox{1.0\columnwidth}{!}{
        \setlength\tabcolsep{8pt}
        \renewcommand\arraystretch{1.0}
\begin{tabular}{lccccccc}
\toprule
\textbf{Model}&\textbf{Metric} & \textbf{MATH500} & \textbf{AIME24} & \textbf{AIME25} & \textbf{HumanEval} & \makecell{\textbf{LiveCodeBench}\\(v6-lite)}  & \makecell{\textbf{LiveBench}\\(reasoning)}\\

\midrule
\makecell[l]{ \cbox{0} Think \cbox{0} DES \\ \cbox{0} Instruct tune} & Acc(\%)   & 79.60 & 23.33 & 13.33  & 87.19 & 20.61 & 41.50 \\
\midrule
\makecell[l]{ \cbox{1} Think \cbox{0} DES \\ \cbox{0} Instruct tune} & Acc(\%)   & 92.40 & 83.33 & \textbf{70.00} & 88.41 & \textbf{37.40} & 63.00  \\
\midrule
\textbf{\makecell[l]{ \cbox{0} Think \cbox{1} DES \\ \cbox{0} Instruct tune}} & Acc(\%)   & 84.80 & 43.33 & 20.00 & 74.39 & 21.37 & 57.50 \\
\midrule
\makecell[l]{ \cbox{0} Think \cbox{0} DES \\ \cbox{1} Instruct tune} & Acc(\%)   & 92.00 & 70.00 & 63.33 & 91.46 & 31.29 & 81.00  \\
\midrule
\textbf{\makecell[l]{ \cbox{0} Think \cbox{1} DES \\ \cbox{1} Instruct tune}} & Acc(\%)   & \textbf{93.20} & \textbf{86.67} & \textbf{70.00} & \textbf{94.51} & 33.59 & \textbf{91.50} \\

\bottomrule
\end{tabular}}
\vspace{-10pt}
\end{table*}

\begin{figure*}[t]
\centering
\includegraphics[width=\textwidth]{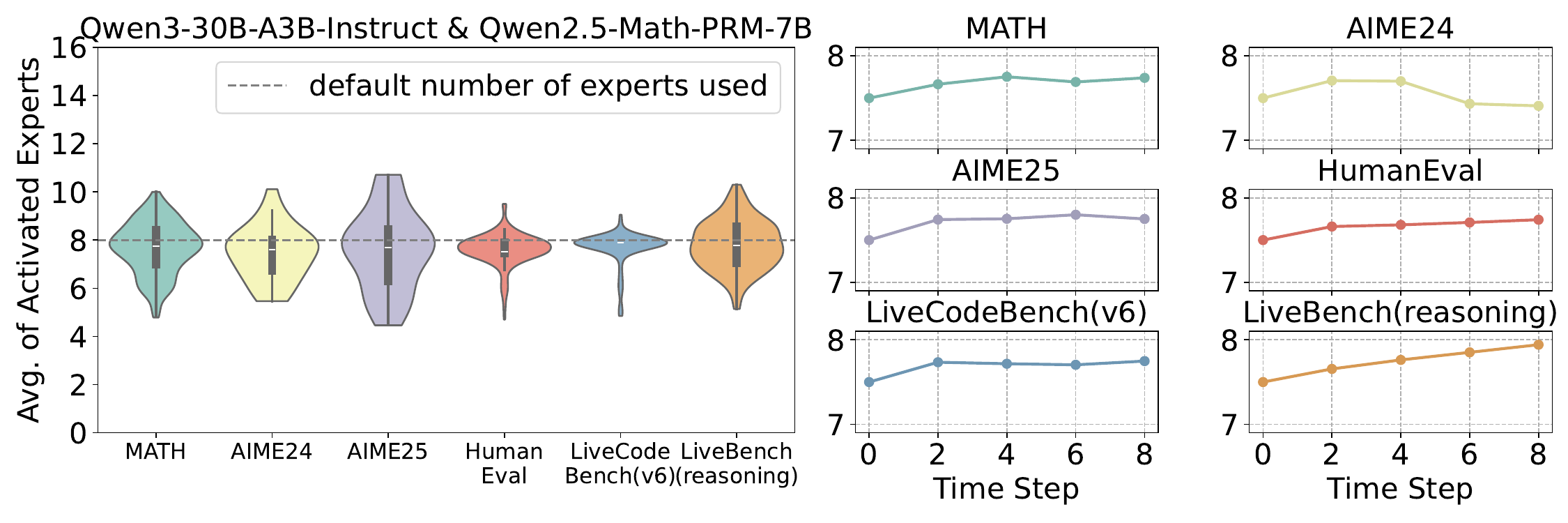}
\caption{Left: Average number of activated experts on various datasets using DES. Right: Average number of activated experts at each timestep of DES on various datasets.}
\label{violin}
\vspace{-15pt}
\end{figure*}
\subsection{Main Result}

\begin{table*}[t]
  \caption{Ablation study on the initial numbers of activated experts when applying DES. Results are obtained on \texttt{Qwen3-30B-A3B-Instruct}, with $N=32$ rollouts per problem.}
  \label{tab:vary_number}
  \centering
  \resizebox{0.8\textwidth}{!}{%
  \begin{tabular}{ccccccccccc}
    \toprule
     & \multicolumn{9}{c}{Initial numbers of activated experts} \\
     \cmidrule(r){2-10}
    \multirow{-2}{*}{Metric} &2-9 & 3-10 & 4-11 & 5-12 & 6-13 & 7-14 & 8-15 &9-16&10-17 \\
    \midrule
    \makecell{Avg.\#Experts} & 5.84  & 6.79 & 7.62  & 8.44 & 9.47 & 10.53 & 11.02 & 11.95 & 13.11\\
    \bottomrule
  \end{tabular}}
  \vspace{-5pt}
\end{table*}
\begin{table*}[t]
\centering
\caption{\texttt{Average number of generated tokens per problem} (\#Gen.Tok$\downarrow$) for different search strategies. Each problem is solved with $N=32$ rollouts.}
\label{tab:computation}
\setlength{\tabcolsep}{4pt}
\resizebox{\textwidth}{!}{%
\begin{tabular}{lccccccc}
\toprule
\textbf{Strategy} & \textbf{Metric} & \textbf{MATH500} & \textbf{AIME24} & \textbf{AIME25} & \textbf{HumanEval} & \makecell{\textbf{LiveCodeBench}\\(v6-lite)}  & \makecell{\textbf{LiveBench}\\(reasoning)} \\
\midrule
\rowcolor{gray!20}
\multicolumn{8}{c}{\texttt{\textbf{Qwen3-30B-A3B-Instruct \& Qwen2.5-Math-PRM-7B}}} \\
\midrule
Best-of-N & \makecell{\#Gen.Tok}  & 32.9k & 165.5k & 190.8k  & 20.3k & 69.9k & 144.2k \\
\midrule
BeamSearch & \makecell{\#Gen.Tok}   & 33.9k & 169.6k & 199.1k & 20.5k & 73.9k & 134.8k \\
\midrule
DVTS & \makecell{\#Gen.Tok}   & 22.7k & 172.5k & 189.5k & 27.4k & 83.5k & 134.8k \\
\midrule
\textbf{DES(Ours)} & \makecell{\#Gen.Tok} & 33.9k & 168.4k & 197.0k & 20.9k & 87.0k & 135.1k \\
\bottomrule
\end{tabular}
}
\vspace{-18pt}
\end{table*}

\textbf{Comparisons to Different Search Strategies.}
To assess the effectiveness of our proposed DES, we conduct comparative experiments against standard TTS search strategies across benchmark datasets. We report results in terms of \texttt{Accuracy}, \texttt{Precision} and \#Gen.Tok as defined in Section~\ref{subsec:setup}. Specifically, Table~\ref{tab:main_table} presents the results for \texttt{Qwen3-30B-A3B-Instruct} and \texttt{Ling-lite-1.5}, using \texttt{Qwen2.5-Math-PRM-7B} as the verifier. Additional results on other model pairs are provided in \textbf{Appendix~\ref{appendix:a}}.
As shown, DES consistently outperforms TTS baselines in both \texttt{Accuracy} and \texttt{Precision}, highlighting its advantage in guiding the model to allocate computational resources toward more effective exploration during search. This leads to a higher proportion of correct answers and overall improved performance.

\textbf{Comparisons to Thinking Mode.}
Recently, some reasoning models (\eg, \texttt{Qwen3-30B-A3B}) have introduced a \textit{thinking mode}, which generates long chains of thought (\ie, a large number of tokens) to analyze and solve problems in detail. Compared to the non-thinking mode, the thinking mode typically incurs substantially higher computational costs. To provide a comprehensive comparison between \textit{thinking mode} and our proposed DES—two approaches that improve model performance by increasing computational expenditure—we conduct experiments on \texttt{Qwen3-30B-A3B} under different modes (see Table~\ref{tab:compare_with_thinkmode}). The results show that DES achieves comparable or better performance than the thinking mode, highlighting its advantage in balancing effectiveness and efficiency.

\textbf{Comparisons across Different Computational Budgets.}
We evaluate how model performance scales with the computational budget $N$ by comparing \texttt{OLMoE-1B-7B-Instruct \& Llama3.1-8B-PRM-Deepseek-Data} on \texttt{GSM8K} and \texttt{MATH500} under different search strategies. As shown in Figure~\ref{search_strategy_comparison}, our method consistently outperforms the baselines across nearly all budgets. Moreover, the performance gap widens as $N$ increases, indicating that our approach raises the upper bound of achievable accuracy and demonstrates strong scalability in resource-rich settings.

\subsection{Ablation Study}
\textbf{Average Number of Activated Experts.}
To ensure that the performance gains of our method are not merely due to activating more experts, we measure the average number of activated experts during the search process. Figure~\ref{violin}(a) illustrates the distribution of the overall average number of activated experts when using \texttt{Qwen3-30B-A3B-Instruct \& Qwen2.5-Math-PRM-7B} (whose default is 8 experts) across multiple datasets, while Figure~\ref{violin}(b) shows the average number at each timestep. The results show that DES does not activate more experts than the default configuration of the policy models. This suggests that the observed performance improvements stem from the effectiveness of the search strategy rather than merely from activating additional experts.

\textbf{Effect of Initial Numbers of Activated Experts.}
For policy models with a default of 8 activated experts, we vary the initial number of activated experts in $[4,5,6,7,8,9,10,11]$ in experiments shown above. Under this setting, we find that the average number of activated experts during the search process does not exceed the default. To further investigate this phenomenon, we conduct an ablation study on different initial values. As shown in Table~\ref{tab:vary_number}, the observed average consistently aligns with the mean of the initial values(\eg, 7.5 for $[4,5,6,7,8,9,10,11]$). This indicates that the average number of activated experts in the search process is largely determined by the initial setting. Moreover, the results suggest that all candidate numbers of activated experts are uniformly explored during search, rather than the model simply favoring larger values.
\begin{table*}[t]
  \caption{Ablation study on Dynamic MoE which introduces \texttt{Exploration on different numbers of activated experts} (Exp.on.Num). We performed on \texttt{MATH500} benchmark using \texttt{OLMoE-1B-7B-Instruct \& Llama3.1-8B-PRM-Deepseek-Data}.}
  \label{exploration}
  \centering
  \resizebox{0.8\textwidth}{!}{%
  \begin{tabular}{ccccccc}
    \toprule
     & & \multicolumn{5}{c}{Computational budget $N$} \\
     \cmidrule(r){3-7}
    \multirow{-2}{*}{Metric} & \multirow{-2}{*}{Strategy} & 8 & 16 & 32 & 64 & 128 \\
    \midrule
    
    \multirow{2}{*}{Acc(\%)}   & BeamSearch w/o Exp.on.Num  & \textbf{21.00}& 23.00&24.60&28.40 & 27.40  \\
                            & BeamSearch w Exp.on.Num &  20.40 & \textbf{24.20}  &  \textbf{25.60} & \textbf{29.60}  &  \textbf{31.00} \\
    \bottomrule
  \end{tabular}}
  \vspace{-10pt}
\end{table*}
\begin{figure*}[t]
\centering
\includegraphics[width=\textwidth]{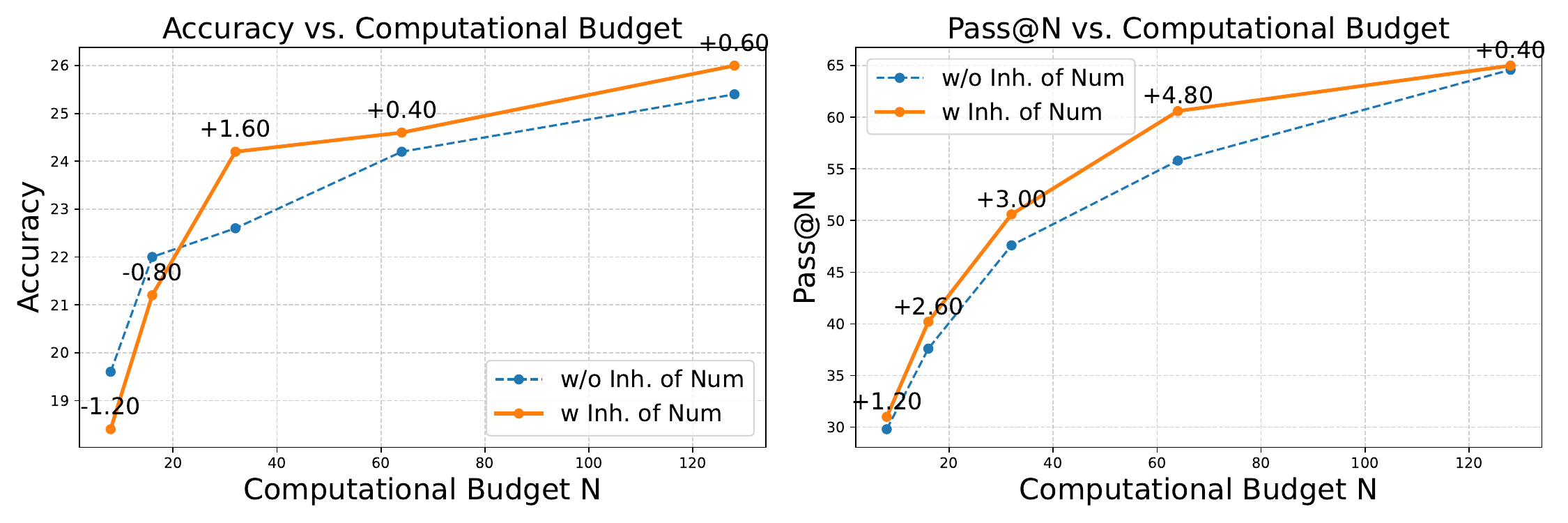}
\vspace{-10pt}
\caption{Ablation study on \texttt{Inheritance of the number of activated experts} (Inh.of.Num) conducted on the \texttt{MATH500} benchmark, using \texttt{OLMoE-1B-7B-Instruct} as the policy model and \texttt{Llama3.1-8B-PRM-Deepseek-Data} as the verifier. The left subfigure reports Accuracy of DES with and without Inh.of.Num, while the right subfigure presents pass@$N$.}
\label{inheritance}
\vspace{-15pt}
\end{figure*}

\textbf{Computation Comparison Between Baselines and DES.}
In the performance comparison experiments, both the baseline strategies and DES generate the same number of rollouts. However, an identical number of rollouts does not strictly imply equal computational cost across search strategies. To enable a more comprehensive comparison, we additionally report the \texttt{average number of generated tokens per problem} (\#Gen.Tok$\downarrow$) for each strategy (see Table~\ref{tab:computation}). Since the search strategies do not alter the model architecture, the computational cost (\eg, FLOPs) per token remains the same; thus, the number of generated tokens is strictly proportional to the overall computation cost. As shown in Table~\ref{tab:computation}, DES incurs a comparable computational cost to the baselines, indicating that the observed performance gains are not attributable to increased computation.

\textbf{Effect of Dynamic MoE.}
To investigate the influence of \texttt{Exploration on different numbers of activated experts} (Exp.on.Num) introduced by Dynamic MoE, we simply compare vanilla \texttt{BeamSearch} and \texttt{BeamSearch} with Dynamic MoE on \texttt{MATH500}. Unlike vanilla \texttt{BeamSearch} activated a fixed number of experts, \texttt{BeamSearch} with Dynamic MoE uniformly varies different numbers of activated experts to generate multiple candidate next reasoning steps at each timestep. As shown in Table~\ref{exploration}, introducing the exploration of different numbers of activated experts leads to a stable performance improvement with different computational budget $N$. 

\textbf{Effect of Experts Configuration Inheritance.}
The \texttt{Inheritance of the activated experts number} is designed to guide computation toward more suitable configurations, thereby improving the effectiveness of exploration during the search process. To assess the impact of \texttt{Inheritance of the activated experts number} (Inh.of.Num), we conduct an ablation study comparing DES with and without this setting. In the variant without Inh.of.Num, different numbers of experts are uniformly explored at each timestep when generating candidates. As shown in Figure~\ref{inheritance}, enabling Inh.of.Num improves not only accuracy but also the pass@$N$ metric, which measures the probability of obtaining a correct answer within $N$ attempts under a fixed computational budget. These results demonstrate that Inh.of.Num effectively steers the search process in a more promising direction and substantially increases the likelihood of generating correct answers.

\section{Limitation}
\label{sec:limitation}

While DES achieves performance improvements, it also shares common limitations with other TTS strategies. First, it requires guidance from an external verifier during the search process, which introduces additional communication overhead. Second, the performance of DES is partly dependent on the quality and reliability of the verifier, as misaligned evaluations can hinder final performance.

\section{Conclusion}
\label{sec:conclusion}

We propose {Dynamic Experts Search} (DES), a novel Test-Time Scaling (TTS) strategy that harnesses the modular structure of MoE models to unlock complementary reasoning capabilities. By dynamically adjusting the number of activated experts during inference, DES introduces expert configuration as a controllable dimension within the search space, enabling capacity-aware exploration.
Extensive experiments across multiple reasoning benchmarks demonstrate the consistent advantage of DES over existing TTS baselines. Beyond performance gains, our work offers a new perspective on Test-Time Scaling by highlighting the untapped flexibility within certain distinctive architectures. We believe DES paves the way for more adaptive and structurally-aware reasoning in large language models.

\bibliography{iclr2026_conference}

\begin{thebibliography}{49}
\providecommand{\natexlab}[1]{#1}
\providecommand{\url}[1]{\texttt{#1}}
\expandafter\ifx\csname urlstyle\endcsname\relax
  \providecommand{\doi}[1]{doi: #1}\else
  \providecommand{\doi}{doi: \begingroup \urlstyle{rm}\Url}\fi

\bibitem[{AI-MO}(2024)]{AIME24}
{AI-MO}.
\newblock Aime 2024, 2024.
\newblock URL \url{https://huggingface.co/datasets/AI-MO/aimo-validation-aime}.

\bibitem[{AI-MO}(2025)]{AIME25}
{AI-MO}.
\newblock Aime 2025, 2025.
\newblock URL \url{https://huggingface.co/datasets/AI-MO/aimo-validation-aime}.

\bibitem[Anthropic(2023)]{Claude}
Anthropic.
\newblock Introducing {Claude}, 2023.
\newblock URL \url{https://www.anthropic.com/index/introducing-claude/}.

\bibitem[Beeching et~al.(2024)Beeching, Tunstall, and Rush]{huggingface2024scaling}
Edward Beeching, Lewis Tunstall, and Sasha Rush.
\newblock Scaling test-time compute with open models, 2024.
\newblock URL \url{https://huggingface.co/spaces/HuggingFaceH4/blogpost-scaling-test-time-compute}.

\bibitem[Brown et~al.(2024)Brown, Juravsky, Ehrlich, Clark, Le, Ré, and Mirhoseini]{brown2024largelanguagemonkeysscaling}
Bradley Brown, Jordan Juravsky, Ryan Ehrlich, Ronald Clark, Quoc~V. Le, Christopher Ré, and Azalia Mirhoseini.
\newblock Large language monkeys: Scaling inference compute with repeated sampling, 2024.

\bibitem[Chen et~al.(2024)Chen, Liao, Li, and Fan]{chen2024alphamath}
Guoxin Chen, Minpeng Liao, Chengxi Li, and Kai Fan.
\newblock Alphamath almost zero: Process supervision without process.
\newblock In \emph{Neural Information Processing Systems}, 2024.

\bibitem[Chen et~al.(2021)Chen, Tworek, Jun, Yuan, de~Oliveira~Pinto, Kaplan, Edwards, Burda, Joseph, Brockman, Ray, Puri, Krueger, Petrov, Khlaaf, Sastry, Mishkin, Chan, Gray, Ryder, Pavlov, Power, Kaiser, Bavarian, Winter, Tillet, Such, Cummings, Plappert, Chantzis, Barnes, Herbert-Voss, Guss, Nichol, Paino, Tezak, Tang, Babuschkin, Balaji, Jain, Saunders, Hesse, Carr, Leike, Achiam, Misra, Morikawa, Radford, Knight, Brundage, Murati, Mayer, Welinder, McGrew, Amodei, McCandlish, Sutskever, and Zaremba]{chen2021evaluating}
Mark Chen, Jerry Tworek, Heewoo Jun, Qiming Yuan, Henrique~Ponde de~Oliveira~Pinto, Jared Kaplan, Harri Edwards, Yuri Burda, Nicholas Joseph, Greg Brockman, Alex Ray, Raul Puri, Gretchen Krueger, Michael Petrov, Heidy Khlaaf, Girish Sastry, Pamela Mishkin, Brooke Chan, Scott Gray, Nick Ryder, Mikhail Pavlov, Alethea Power, Lukasz Kaiser, Mohammad Bavarian, Clemens Winter, Philippe Tillet, Felipe~Petroski Such, Dave Cummings, Matthias Plappert, Fotios Chantzis, Elizabeth Barnes, Ariel Herbert-Voss, William~Hebgen Guss, Alex Nichol, Alex Paino, Nikolas Tezak, Jie Tang, Igor Babuschkin, Suchir Balaji, Shantanu Jain, William Saunders, Christopher Hesse, Andrew~N. Carr, Jan Leike, Josh Achiam, Vedant Misra, Evan Morikawa, Alec Radford, Matthew Knight, Miles Brundage, Mira Murati, Katie Mayer, Peter Welinder, Bob McGrew, Dario Amodei, Sam McCandlish, Ilya Sutskever, and Wojciech Zaremba.
\newblock Evaluating large language models trained on code, 2021.

\bibitem[Cobbe et~al.(2021)Cobbe, Kosaraju, Bavarian, Chen, Jun, Kaiser, Plappert, Tworek, Hilton, Nakano, et~al.]{GSM8K}
Karl Cobbe, Vineet Kosaraju, Mohammad Bavarian, Mark Chen, Heewoo Jun, Lukasz Kaiser, Matthias Plappert, Jerry Tworek, Jacob Hilton, Reiichiro Nakano, et~al.
\newblock Training verifiers to solve math word problems.
\newblock \emph{arXiv preprint arXiv:2110.14168}, 2021.

\bibitem[DeepSeek-AI(2024)]{deepseekv2}
DeepSeek-AI.
\newblock Deepseek-v2: A strong, economical, and efficient mixture-of-experts language model, 2024.

\bibitem[Dubey et~al.(2024)Dubey, Jauhri, Pandey, Kadian, Al-Dahle, Letman, Mathur, Schelten, Yang, Fan, Goyal, Hartshorn, Yang, Mitra, Sravankumar, Korenev, Hinsvark, Rao, Zhang, Rodriguez, Gregerson, Spataru, Rozière, Biron, Tang, Chern, Caucheteux, Nayak, Bi, Marra, McConnell, Keller, Touret, Wu, Wong, Ferrer, Nikolaidis, Allonsius, Song, Pintz, Livshits, Esiobu, Choudhary, Mahajan, Garcia-Olano, Perino, Hupkes, Lakomkin, AlBadawy, Lobanova, Dinan, Smith, Radenovic, Zhang, Synnaeve, Lee, Anderson, Nail, Mialon, Pang, Cucurell, Nguyen, Korevaar, Xu, Touvron, Zarov, Ibarra, Kloumann, Misra, Evtimov, Copet, Lee, Geffert, Vranes, Park, Mahadeokar, Shah, van~der Linde, Billock, Hong, Lee, Fu, Chi, Huang, Liu, Wang, Yu, Bitton, Spisak, Park, Rocca, Johnstun, Saxe, Jia, Alwala, Upasani, Plawiak, Li, Heafield, Stone, and et~al.]{DBLP:journals/corr/abs-2407-21783}
Abhimanyu Dubey, Abhinav Jauhri, Abhinav Pandey, Abhishek Kadian, Ahmad Al-Dahle, Aiesha Letman, Akhil Mathur, Alan Schelten, Amy Yang, Angela Fan, Anirudh Goyal, Anthony Hartshorn, Aobo Yang, Archi Mitra, Archie Sravankumar, Artem Korenev, Arthur Hinsvark, Arun Rao, Aston Zhang, Aurélien Rodriguez, Austen Gregerson, Ava Spataru, Baptiste Rozière, Bethany Biron, Binh Tang, Bobbie Chern, Charlotte Caucheteux, Chaya Nayak, Chloe Bi, Chris Marra, Chris McConnell, Christian Keller, Christophe Touret, Chunyang Wu, Corinne Wong, Cristian~Canton Ferrer, Cyrus Nikolaidis, Damien Allonsius, Daniel Song, Danielle Pintz, Danny Livshits, David Esiobu, Dhruv Choudhary, Dhruv Mahajan, Diego Garcia-Olano, Diego Perino, Dieuwke Hupkes, Egor Lakomkin, Ehab AlBadawy, Elina Lobanova, Emily Dinan, Eric~Michael Smith, Filip Radenovic, Frank Zhang, Gabriel Synnaeve, Gabrielle Lee, Georgia~Lewis Anderson, Graeme Nail, Grégoire Mialon, Guan Pang, Guillem Cucurell, Hailey Nguyen, Hannah Korevaar, Hu~Xu, Hugo Touvron, Iliyan Zarov,
  Imanol~Arrieta Ibarra, Isabel~M. Kloumann, Ishan Misra, Ivan Evtimov, Jade Copet, Jaewon Lee, Jan Geffert, Jana Vranes, Jason Park, Jay Mahadeokar, Jeet Shah, Jelmer van~der Linde, Jennifer Billock, Jenny Hong, Jenya Lee, Jeremy Fu, Jianfeng Chi, Jianyu Huang, Jiawen Liu, Jie Wang, Jiecao Yu, Joanna Bitton, Joe Spisak, Jongsoo Park, Joseph Rocca, Joshua Johnstun, Joshua Saxe, Junteng Jia, Kalyan~Vasuden Alwala, Kartikeya Upasani, Kate Plawiak, Ke~Li, Kenneth Heafield, Kevin Stone, and et~al.
\newblock The llama 3 herd of models.
\newblock \emph{CoRR}, abs/2407.21783, 2024.

\bibitem[Fedus et~al.(2022)Fedus, Zoph, and Shazeer]{SwitchTransformers}
William Fedus, Barret Zoph, and Noam Shazeer.
\newblock Switch transformers: scaling to trillion parameter models with simple and efficient sparsity.
\newblock \emph{J. Mach. Learn. Res.}, 23, 2022.

\bibitem[Guo et~al.(2025)Guo, Yang, Zhang, Song, Zhang, Xu, Zhu, Ma, Wang, Bi, et~al.]{guo2025deepseek}
Daya Guo, Dejian Yang, Haowei Zhang, Junxiao Song, Ruoyu Zhang, Runxin Xu, Qihao Zhu, Shirong Ma, Peiyi Wang, Xiao Bi, et~al.
\newblock Deepseek-r1: Incentivizing reasoning capability in llms via reinforcement learning.
\newblock \emph{arXiv preprint arXiv:2501.12948}, 2025.

\bibitem[Huang et~al.(2025)Huang, Yang, Liu, Xu, Li, Liu, Yin, Li, Ren, and Barsoum]{huang2025jakiroboostingspeculativedecoding}
Haiduo Huang, Fuwei Yang, Zhenhua Liu, Yixing Xu, Jinze Li, Yang Liu, Xuanwu Yin, Dong Li, Pengju Ren, and Emad Barsoum.
\newblock Jakiro: Boosting speculative decoding with decoupled multi-head via moe, 2025.

\bibitem[Huang et~al.(2024)Huang, An, Zhuang, Tao, Zhang, Jin, Xu, Chen, Huang, and Feng]{DBLP:journals/corr/abs-2403-07652}
Quzhe Huang, Zhenwei An, Nan Zhuang, Mingxu Tao, Chen Zhang, Yang Jin, Kun Xu, Liwei Chen, Songfang Huang, and Yansong Feng.
\newblock Harder tasks need more experts: Dynamic routing in moe models.
\newblock \emph{CoRR}, abs/2403.07652, 2024.

\bibitem[Hurst et~al.(2024)Hurst, Lerer, Goucher, Perelman, Ramesh, Clark, Ostrow, Welihinda, Hayes, Radford, et~al.]{GPT-4o}
Aaron Hurst, Adam Lerer, Adam~P Goucher, Adam Perelman, Aditya Ramesh, Aidan Clark, AJ~Ostrow, Akila Welihinda, Alan Hayes, Alec Radford, et~al.
\newblock Gpt-4o system card.
\newblock \emph{arXiv preprint arXiv:2410.21276}, 2024.

\bibitem[Jacobs et~al.(1991)Jacobs, Jordan, Nowlan, and Hinton]{Jacobs_Jordan_Nowlan_Hinton_1991}
Robert~A. Jacobs, Michael~I. Jordan, Steven~J. Nowlan, and Geoffrey~E. Hinton.
\newblock Adaptive mixtures of local experts.
\newblock \emph{Neural Computation}, pp.\  79–87, 1991.

\bibitem[Jain et~al.(2025)Jain, Han, Gu, Li, Yan, Zhang, Wang, Solar-Lezama, Sen, and Stoica]{jain2025livecodebench}
Naman Jain, King Han, Alex Gu, Wen-Ding Li, Fanjia Yan, Tianjun Zhang, Sida Wang, Armando Solar-Lezama, Koushik Sen, and Ion Stoica.
\newblock Livecodebench: Holistic and contamination free evaluation of large language models for code.
\newblock In \emph{International Conference on Learning Representations}, 2025.

\bibitem[Jin et~al.(2025)Jin, Zhu, Yuan, and YAN]{jin2025moe}
Peng Jin, Bo~Zhu, Li~Yuan, and Shuicheng YAN.
\newblock Moe++: Accelerating mixture-of-experts methods with zero-computation experts.
\newblock In \emph{International Conference on Learning Representations}, 2025.

\bibitem[Kang et~al.(2024)Kang, Li, Chen, Kazemi, Sun, Chen, Li, He, He, Wen, et~al.]{MindStar}
Jikun Kang, Xin~Zhe Li, Xi~Chen, Amirreza Kazemi, Qianyi Sun, Boxing Chen, Dong Li, Xu~He, Quan He, Feng Wen, et~al.
\newblock {MindStar}: Enhancing math reasoning in pre-trained llms at inference time.
\newblock \emph{arXiv preprint arXiv:2405.16265}, 2024.

\bibitem[Kwon et~al.(2023)Kwon, Li, Zhuang, Sheng, Zheng, Yu, Gonzalez, Zhang, and Stoica]{kwon2023efficient}
Woosuk Kwon, Zhuohan Li, Siyuan Zhuang, Ying Sheng, Lianmin Zheng, Cody~Hao Yu, Joseph~E. Gonzalez, Hao Zhang, and Ion Stoica.
\newblock Efficient memory management for large language model serving with pagedattention.
\newblock In \emph{Proceedings of the ACM SIGOPS 29th Symposium on Operating Systems Principles}, 2023.

\bibitem[Lepikhin et~al.(2021)Lepikhin, Lee, Xu, Chen, Firat, Huang, Krikun, Shazeer, and Chen]{lepikhin2021gshard}
Dmitry Lepikhin, HyoukJoong Lee, Yuanzhong Xu, Dehao Chen, Orhan Firat, Yanping Huang, Maxim Krikun, Noam Shazeer, and Zhifeng Chen.
\newblock {\{}GS{\}}hard: Scaling giant models with conditional computation and automatic sharding.
\newblock In \emph{International Conference on Learning Representations}, 2021.

\bibitem[Lewis et~al.(2021)Lewis, Bhosale, Dettmers, Goyal, and Zettlemoyer]{lewis2021baselayerssimplifyingtraining}
Mike Lewis, Shruti Bhosale, Tim Dettmers, Naman Goyal, and Luke Zettlemoyer.
\newblock Base layers: Simplifying training of large, sparse models, 2021.

\bibitem[Li et~al.(2023)Li, Su, Yang, Jiang, Wang, and Xu]{li2023adaptive}
Jiamin Li, Qiang Su, Yitao Yang, Yimin Jiang, Cong Wang, and Hong Xu.
\newblock Adaptive gating in mixture-of-experts based language models.
\newblock In \emph{Empirical Methods in Natural Language Processing}, 2023.

\bibitem[Lightman et~al.(2024)Lightman, Kosaraju, Burda, Edwards, Baker, Lee, Leike, Schulman, Sutskever, and Cobbe]{PRM800K}
Hunter Lightman, Vineet Kosaraju, Yuri Burda, Harrison Edwards, Bowen Baker, Teddy Lee, Jan Leike, John Schulman, Ilya Sutskever, and Karl Cobbe.
\newblock Let's verify step by step.
\newblock In \emph{International Conference on Learning Representations}, 2024.

\bibitem[Ling(2025)]{ling}
Ling.
\newblock Every flop counts: Scaling a 300b mixture-of-experts ling llm without premium gpus.
\newblock \emph{arXiv preprint arXiv:2503.05139}, 2025.

\bibitem[Liu et~al.(2025)Liu, Gao, Zhao, Zhang, Li, Qi, Ouyang, and Zhou]{liu2025can}
Runze Liu, Junqi Gao, Jian Zhao, Kaiyan Zhang, Xiu Li, Biqing Qi, Wanli Ouyang, and Bowen Zhou.
\newblock Can 1b llm surpass 405b llm? rethinking compute-optimal test-time scaling.
\newblock \emph{arXiv preprint arXiv:2502.06703}, 2025.

\bibitem[Muennighoff et~al.(2025)Muennighoff, Soldaini, Groeneveld, Lo, Morrison, Min, Shi, Walsh, Tafjord, Lambert, Gu, Arora, Bhagia, Schwenk, Wadden, Wettig, Hui, Dettmers, Kiela, Farhadi, Smith, Koh, Singh, and Hajishirzi]{muennighoff2025olmoe}
Niklas Muennighoff, Luca Soldaini, Dirk Groeneveld, Kyle Lo, Jacob Morrison, Sewon Min, Weijia Shi, Evan~Pete Walsh, Oyvind Tafjord, Nathan Lambert, Yuling Gu, Shane Arora, Akshita Bhagia, Dustin Schwenk, David Wadden, Alexander Wettig, Binyuan Hui, Tim Dettmers, Douwe Kiela, Ali Farhadi, Noah~A. Smith, Pang~Wei Koh, Amanpreet Singh, and Hannaneh Hajishirzi.
\newblock {OLM}oe: Open mixture-of-experts language models.
\newblock In \emph{International Conference on Learning Representations}, 2025.

\bibitem[OpenAI(2024)]{o1}
OpenAI.
\newblock Learning to reason with llms, 2024.
\newblock URL \url{https://openai.com/index/learning-to-reason-with-llms/}.

\bibitem[Qi et~al.(2025)Qi, MA, Xu, Zhang, Yang, and Yang]{qi2025mutual}
Zhenting Qi, Mingyuan MA, Jiahang Xu, Li~Lyna Zhang, Fan Yang, and Mao Yang.
\newblock Mutual reasoning makes smaller {LLM}s stronger problem-solver.
\newblock In \emph{International Conference on Learning Representations}, 2025.

\bibitem[Qu et~al.(2024)Qu, Zhang, Garg, and Kumar]{RISE}
Yuxiao Qu, Tianjun Zhang, Naman Garg, and Aviral Kumar.
\newblock Recursive introspection: Teaching language model agents how to self-improve.
\newblock In \emph{Advances in Neural Information Processing Systems}, 2024.

\bibitem[Qwen(2025)]{qwen3technicalreport}
Qwen.
\newblock Qwen3 technical report, 2025.
\newblock URL \url{https://arxiv.org/abs/2505.09388}.

\bibitem[Roller et~al.(2021)Roller, Sukhbaatar, Szlam, and Weston]{Roller_Sukhbaatar_Szlam_Weston_2021}
Stephen Roller, Sainbayar Sukhbaatar, Arthur Szlam, and Jason Weston.
\newblock Hash layers for large sparse models.
\newblock \emph{Neural Information Processing Systems}, 2021.

\bibitem[Saha et~al.(2021)Saha, Dubey, and Hovy]{saha-etal-2021-svamp}
Amrita Saha, Mohnish Dubey, and Eduard Hovy.
\newblock Svamp: A dataset for evaluating verbal reasoning in math word problems.
\newblock In \emph{Findings of the Association for Computational Linguistics: EMNLP 2021}, 2021.

\bibitem[Shazeer et~al.(2017)Shazeer, Mirhoseini, Maziarz, Davis, Le, Hinton, and Dean]{shazeer2017}
Noam Shazeer, *Azalia Mirhoseini, *Krzysztof Maziarz, Andy Davis, Quoc Le, Geoffrey Hinton, and Jeff Dean.
\newblock Outrageously large neural networks: The sparsely-gated mixture-of-experts layer.
\newblock In \emph{International Conference on Learning Representations}, 2017.

\bibitem[Snell et~al.(2024)Snell, Lee, Xu, and Kumar]{snell2024scaling}
Charlie Snell, Jaehoon Lee, Kelvin Xu, and Aviral Kumar.
\newblock Scaling llm test-time compute optimally can be more effective than scaling model parameters.
\newblock \emph{arXiv preprint arXiv:2408.03314}, 2024.

\bibitem[Wan et~al.(2024)Wan, Feng, Wen, Mcaleer, Wen, Zhang, and Wang]{wan2024alphazero}
Ziyu Wan, Xidong Feng, Muning Wen, Stephen~Marcus Mcaleer, Ying Wen, Weinan Zhang, and Jun Wang.
\newblock {A}lpha{Z}ero-like tree-search can guide large language model decoding and training.
\newblock In \emph{International Conference on Machine Learning}, 2024.

\bibitem[Wang et~al.(2024)Wang, Li, Shao, Xu, Dai, Li, Chen, Wu, and Sui]{wang2024math}
Peiyi Wang, Lei Li, Zhihong Shao, Runxin Xu, Damai Dai, Yifei Li, Deli Chen, Yu~Wu, and Zhifang Sui.
\newblock Math-shepherd: Verify and reinforce llms step-by-step without human annotations.
\newblock In \emph{Proceedings of the 62nd Annual Meeting of the Association for Computational Linguistics (Volume 1: Long Papers)}, pp.\  9426--9439, 2024.

\bibitem[Wang et~al.(2023)Wang, Wei, Schuurmans, Le, Chi, Narang, Chowdhery, and Zhou]{Self-Consistency}
Xuezhi Wang, Jason Wei, Dale Schuurmans, Quoc~V Le, Ed~H. Chi, Sharan Narang, Aakanksha Chowdhery, and Denny Zhou.
\newblock Self-consistency improves chain of thought reasoning in language models.
\newblock In \emph{International Conference on Learning Representations}, 2023.

\bibitem[Wang et~al.(2025)Wang, Zhu, and Chen]{wang2025remoe}
Ziteng Wang, Jun Zhu, and Jianfei Chen.
\newblock Remoe: Fully differentiable mixture-of-experts with re{LU} routing.
\newblock In \emph{International Conference on Learning Representations}, 2025.

\bibitem[Wei et~al.(2022)Wei, Wang, Schuurmans, Bosma, brian ichter, Xia, Chi, Le, and Zhou]{wei2022chain}
Jason Wei, Xuezhi Wang, Dale Schuurmans, Maarten Bosma, brian ichter, Fei Xia, Ed~H. Chi, Quoc~V Le, and Denny Zhou.
\newblock Chain of thought prompting elicits reasoning in large language models.
\newblock In \emph{Advances in Neural Information Processing Systems}, 2022.

\bibitem[White et~al.(2025)White, Dooley, Roberts, Pal, Feuer, Jain, Shwartz-Ziv, Jain, Saifullah, Dey, Shubh-Agrawal, Sandha, Naidu, Hegde, LeCun, Goldstein, Neiswanger, and Goldblum]{white2025livebench}
Colin White, Samuel Dooley, Manley Roberts, Arka Pal, Benjamin Feuer, Siddhartha Jain, Ravid Shwartz-Ziv, Neel Jain, Khalid Saifullah, Sreemanti Dey, Shubh-Agrawal, Sandeep~Singh Sandha, Siddartha~Venkat Naidu, Chinmay Hegde, Yann LeCun, Tom Goldstein, Willie Neiswanger, and Micah Goldblum.
\newblock Livebench: A challenging, contamination-limited {LLM} benchmark.
\newblock In \emph{International Conference on Learning Representations}, 2025.

\bibitem[Wolf et~al.(2020)Wolf, Debut, Sanh, Chaumond, Delangue, Moi, Cistac, Rault, Louf, Funtowicz, Davison, Shleifer, von Platen, Ma, Jernite, Plu, Xu, Scao, Gugger, Drame, Lhoest, and Rush]{wolf-etal-2020-transformers}
Thomas Wolf, Lysandre Debut, Victor Sanh, Julien Chaumond, Clement Delangue, Anthony Moi, Pierric Cistac, Tim Rault, Rémi Louf, Morgan Funtowicz, Joe Davison, Sam Shleifer, Patrick von Platen, Clara Ma, Yacine Jernite, Julien Plu, Canwen Xu, Teven~Le Scao, Sylvain Gugger, Mariama Drame, Quentin Lhoest, and Alexander~M. Rush.
\newblock Transformers: State-of-the-art natural language processing.
\newblock In \emph{Proceedings of the 2020 Conference on Empirical Methods in Natural Language Processing: System Demonstrations}, 2020.

\bibitem[Wu et~al.(2024)Wu, Sun, Li, Welleck, and Yang]{wu2024scaling}
Yangzhen Wu, Zhiqing Sun, Shanda Li, Sean Welleck, and Yiming Yang.
\newblock Scaling inference computation: Compute-optimal inference for problem-solving with language models.
\newblock In \emph{Advances in Neural Information Processing Systems}, 2024.

\bibitem[Xie et~al.(2023)Xie, Kawaguchi, Zhao, Zhao, Kan, He, and Xie]{xie2024self}
Yuxi Xie, Kenji Kawaguchi, Yiran Zhao, James~Xu Zhao, Min-Yen Kan, Junxian He, and Michael Xie.
\newblock Self-evaluation guided beam search for reasoning.
\newblock In \emph{Advances in Neural Information Processing Systems}, 2023.

\bibitem[Xu et~al.(2025)Xu, Yan, Ma, Zhao, Liu, Lin, and Wu]{xu2025phidecodingadaptiveforesightsampling}
Fangzhi Xu, Hang Yan, Chang Ma, Haiteng Zhao, Jun Liu, Qika Lin, and Zhiyong Wu.
\newblock $\phi$-decoding: Adaptive foresight sampling for balanced inference-time exploration and exploitation, 2025.

\bibitem[Yao et~al.(2023)Yao, Yu, Zhao, Shafran, Griffiths, Cao, and Narasimhan]{ToT}
Shunyu Yao, Dian Yu, Jeffrey Zhao, Izhak Shafran, Tom Griffiths, Yuan Cao, and Karthik Narasimhan.
\newblock Tree of thoughts: Deliberate problem solving with large language models.
\newblock In \emph{Advances in Neural Information Processing Systems}, 2023.

\bibitem[Zeng et~al.(2024)Zeng, Miao, Gao, Zhang, and Deng]{anonymous2024adamoe}
Zihao Zeng, Yibo Miao, Hongcheng Gao, Hao Zhang, and Zhijie Deng.
\newblock Adamoe: Token-adaptive routing with null experts for mixture-of-experts language models.
\newblock In \emph{Submitted to ACL Rolling Review - June 2024}, 2024.
\newblock under review.

\bibitem[Zhang et~al.(2025)Zhang, Zheng, Wu, Zhang, Lin, Yu, Liu, Zhou, and Lin]{anonymous2025the}
Zhenru Zhang, Chujie Zheng, Yangzhen Wu, Beichen Zhang, Runji Lin, Bowen Yu, Dayiheng Liu, Jingren Zhou, and Junyang Lin.
\newblock The lessons of developing process reward models in mathematical reasoning.
\newblock In \emph{Submitted to ACL Rolling Review - February 2025}, 2025.
\newblock under review.

\bibitem[Zhou et~al.(2022)Zhou, Lei, Liu, Du, Huang, Zhao, Dai, Chen, Le, and Laudon]{zhou2022mixtureofexperts}
Yanqi Zhou, Tao Lei, Hanxiao Liu, Nan Du, Yanping Huang, Vincent~Y Zhao, Andrew~M. Dai, Zhifeng Chen, Quoc~V Le, and James Laudon.
\newblock Mixture-of-experts with expert choice routing.
\newblock In \emph{Advances in Neural Information Processing Systems}, 2022.

\end{thebibliography}
\bibliographystyle{iclr2026_conference}
\newpage
\appendix
\section*{SUMMARY OF THE APPENDIX\centering}

This appendix contains additional details for the ICLR 2026 submission, titled "Enhancing LLM Reasoning in Test-Time Scaling via Dynamic Experts Search". The appendix is organized as follows:

\begin{itemize}
    \item Section~\ref{appendix:a} reports additional experiments results and analysis.
    \item Section~\ref{appendix:b} describes the broader impact of Dynamic Experts Search.
    \item Section~\ref{appendix:c} introduces details of the implementation.
    \item Section~\ref{appendix:d} shows the licenses of datasets, codes and models used in this paper.
    \item Section~\ref{appendix:e} claims the use of large language models (LLMs).
\end{itemize}

\section{Additional Experiments and Analysis}
\label{appendix:a}

\subsection{Experiments using other models}

We further evaluate \texttt{DeepSeek-V2-Lite-Chat} and \texttt{OLMoE-1B-7B-Instruct} on benchmarks from different domains, using \texttt{Qwen2.5-Math-PRM-7B} as the verifier. Due to their relatively small sizes, these two models struggle with challenging benchmarks such as \texttt{AIME}~\cite{AIME24,AIME25}. Therefore, we instead adopt two alternative mathematical reasoning benchmarks (SVAMP~\cite{saha-etal-2021-svamp} and GSM8K~\cite{GSM8K}) for evaluation. As shown in Table~\ref{appendix:main_table}, our method substantially outperforms other search strategies. These results confirm that DES effectively steers the search process toward correct solutions by improving the likelihood of retrieving accurate answers, while also underscoring its generalizability across diverse architectures and model scales.

\begin{table*}[h]
\centering
\caption{\texttt{Accuracy} (Acc $\uparrow$) and \texttt{Precision} (Prec $\uparrow$) of different strategies on benchmarks using \textbf{\texttt{Qwen2.5-Math-PRM-7B}} as policy model. For the implementation of all search methods, we generated $N=64$ rollouts for each problem.}
\label{appendix:main_table}
\setlength{\tabcolsep}{4pt}
\resizebox{\textwidth}{!}{%
\begin{tabular}{lccccccc}
\toprule
\textbf{Strategy} & \textbf{Metric} & \textbf{SVAMP} & \textbf{GSM8K} & \textbf{MATH500} & \textbf{HumanEval} & \makecell{\textbf{LiveCodeBench}\\(v6-lite)}  & \makecell{\textbf{LiveBench}\\(reasoning)} \\
\midrule
\rowcolor{gray!20}
\multicolumn{8}{c}{\texttt{\textbf{DeepSeek-V2-Lite-Chat}}} \\
\midrule
\multirow{2}{*}{Best-of-N}  
        & Acc(\%)    & 85.00 & 68.80 & 30.40  & 40.24 & 10.69 & 13.50 \\
        & Prec(\%)   & 45.13 & 35.14 & 14.79 & 40.37 & 8.14 & 9.45 \\
\midrule
\multirow{2}{*}{BeamSearch} 
        & Acc(\%)    & 83.00 & 69.39 & 35.80 & 41.46 & 11.45 & 13.00 \\
        & Prec(\%)   & 53.28 & 46.17 & 24.58 & 44.61 & 8.36 & 8.71 \\
\midrule
\multirow{2}{*}{DVTS}    
        & Acc(\%)    & 86.00 & 72.60 & 34.00 & 30.49 & 12.21 & 12.00 \\
        & Prec(\%)   & 53.45 & 42.20 & 19.19 & 40.13 & \textbf{8.91} & 8.97 \\
\midrule
\multirow{2}{*}{\textbf{DES(Ours)}} 
        & Acc(\%)    & \textbf{87.67} & \textbf{82.80} & \textbf{41.80} & \textbf{47.56} & \textbf{12.98} & \textbf{16.50} \\
        & Prec(\%)   & \textbf{54.49} & \textbf{49.74} & \textbf{27.01} & \textbf{46.06} & 8.60 & \textbf{10.00} \\
\midrule
\rowcolor{gray!20}
\multicolumn{8}{c}{\texttt{\textbf{OLMoE-1B-7B-Instruct}}} \\
\midrule
\multirow{2}{*}{Best-of-N}  
        & Acc(\%)    & 74.00 & 57.60 & 17.00 & 34.76 & 4.58 & 10.50 \\
        & Prec(\%)   & 33.58 & 22.73 & 7.96 & 27.36 & 2.14 & 6.74 \\
\midrule
\multirow{2}{*}{BeamSearch} 
        & Acc(\%)    & \textbf{84.67} & 74.20 & 25.00 & 34.76 & 4.58 & 10.50 \\
        & Prec(\%)   & 57.51 & 48.74 & 16.62 & 27.36 & \textbf{2.16} & 6.98 \\
\midrule
\multirow{2}{*}{DVTS}    
        & Acc(\%)    & 80.67 & 70.00 & 23.78 & 33.54 & 5.34 & 11.50 \\
        & Prec(\%)   & 50.30 & 37.64 & 12.33 & 27.36 & 1.88 & 7.08 \\
\midrule
\multirow{2}{*}{\textbf{DES(Ours)}} 
        & Acc(\%)    & 83.67 & \textbf{75.40} & \textbf{27.00} & 34.76 & \textbf{6.10} & \textbf{11.50} \\
        & Prec(\%)   & \textbf{60.47} & \textbf{52.86} & \textbf{17.62} & 26.91 & 1.92 & 7.86 \\
\bottomrule
\end{tabular}
}
\vspace{-5pt}
\end{table*}

\subsection{Additional ablation study}

To illustrate the generalizability of DES across different verifier, we also select another alternative process reward model (\texttt{Llama3.1-8B-PRM-Deepseek-Data}) as verifier and evaluate the baseline and DES. As shown in Table~\ref{appendix:ablation_table}, when using a different verifier, DES still outperforms the baseline search method, indicating the improvements brought by DES are not specific to a particular verifier.

\begin{table*}[h]
\centering
\caption{\texttt{Accuracy} (Acc $\uparrow$) and \texttt{Precision} (Prec $\uparrow$) of different strategies on benchmarks using \textbf{\texttt{Llama3.1-8B-PRM-Deepseek-Data}} as policy model. For the implementation of all search methods, we generated $N=64$ rollouts for each problem.}
\label{appendix:ablation_table}
\setlength{\tabcolsep}{4pt}
\resizebox{\textwidth}{!}{%
\begin{tabular}{lccccccc}
\toprule
\textbf{Strategy} & \textbf{Metric} & \textbf{SVAMP} & \textbf{GSM8K} & \textbf{MATH500} & \textbf{HumanEval} & \makecell{\textbf{LiveCodeBench}\\(v6-lite)}  & \makecell{\textbf{LiveBench}\\(reasoning)} \\
\midrule
\rowcolor{gray!20}
\multicolumn{8}{c}{\texttt{\textbf{DeepSeek-V2-Lite-Chat}}} \\
\midrule
\multirow{2}{*}{Best-of-N}  
        & Acc(\%)    & 85.33 & 68.80& 41.00  & 45.12 & 12.21 & 12.00 \\
        & Prec(\%)   & 46.06 & 34.83 & 14.86 & 40.61 & 8.52 & 9.25 \\
\midrule
\multirow{2}{*}{BeamSearch} 
        & Acc(\%)    & 86.33 & 67.60 & 46.60 & 40.85 & \textbf{13.74} & 14.00 \\
        & Prec(\%)   & 57.12 & 44.13 & 27.61 & 40.26 & \textbf{8.53} & 9.85 \\
\midrule
\multirow{2}{*}{DVTS}      
        & Acc(\%)    & 89.33 & 71.00 & 42.40 & 42.68 & 12.21 & 12.50 \\
        & Prec(\%)   & 54.46 & 43.23 & 22.59 & 40.27 & 8.48 & 9.96 \\
\midrule
\multirow{2}{*}{\textbf{DES(Ours)}} 
        & Acc(\%)    & \textbf{91.67} & \textbf{87.60} & \textbf{48.00}  & \textbf{55.49} & 11.45 & \textbf{16.50} \\
        & Prec(\%)   & \textbf{61.47} & \textbf{54.89} & \textbf{28.67} & \textbf{40.90} & 8.43 & \textbf{10.29} \\
\midrule
\rowcolor{gray!20}
\multicolumn{8}{c}{\texttt{\textbf{OLMoE-1B-7B-Instruct}}} \\
\midrule
\multirow{2}{*}{Best-of-N}  
        & Acc(\%)    & 71.67 & 57.60 & 17.40 & 45.73 & \textbf{8.40} & 12.50 \\
        & Prec(\%)   & 32.89 & 22.48 & 7.76 & 27.36 & \textbf{2.16} & 6.14 \\
\midrule
\multirow{2}{*}{BeamSearch} 
        & Acc(\%)    & 86.33 & 76.00 & 27.40 & 45.73 & 6.87 & 12.00 \\
        & Prec(\%)   & 60.00 & 49.87 & 18.67 & 26.91 & 1.93 & 5.58 \\
\midrule
\multirow{2}{*}{DVTS}      
        & Acc(\%)    & 84.00 & 72.60 & 25.20 & 45.73 & 7.63 & 11.50 \\
        & Prec(\%)   & 50.31 & 38.93 & 13.99 & \textbf{27.36} & 2.09 & 6.56 \\
\midrule
\multirow{2}{*}{\textbf{DES(Ours)}} 
        & Acc(\%)    & \textbf{87.00} & \textbf{79.60} & \textbf{29.40} & \textbf{45.73} & 6.10 & \textbf{12.50} \\
        & Prec(\%)   & \textbf{60.85} & \textbf{52.75} & \textbf{19.49} & 26.91 & 1.88 & 6.98 \\
\bottomrule
\end{tabular}
}
\vspace{-5pt}
\end{table*}

\subsection{Why does DES work?}

\textbf{Dynamic MoE Effectively Enhances the Probability of Reaching Correct Answers.}
To demonstrate the effectiveness of Dynamic MoE, we isolate the impact of exploring different numbers of activated experts (denoted as Exp.on.Num) within the beam search framework in ablation study. Specifically, at each step of beam search, we uniformly allocate rollouts across varying numbers of activated experts (\eg, when the total number of rollouts is 128, we generate $\frac{128}{8} = 16$ rollouts for each expert count in $\{4, 5, 6, 7, 8, 9, 10, 11\}$), without incorporating Diversity-Aware Selection or Experts Configuration Inheritance. Compared to vanilla beam search, this strategy expands the search space and makes it possible to reach an appropriate expert configuration capable of generating the correct answer. As a result, the proportion of correct answers among the generated candidates is significantly improved.

\textbf{Experts Configuration Inheritance Avoids Inefficient Computation.}
We adopt Experts Configuration Inheritance to maintain consistency in the expert configuration across steps when generating a complete answer. At each step, inappropriate expert configurations tend to produce low-scoring responses that are subsequently discarded during the search process. Those suboptimal configurations will be filtered out naturally by eliminating low-scoring responses at each step. As a result, in the later stages of search, only effective expert configurations are retained, concentrating computation on more promising candidates and thereby increasing the likelihood of generating the correct answer. In our ablation study, we report the pass@$N$ metric to demonstrate that the final candidate set has a higher probability of containing the correct answer.

\section{Broader Impact}
\label{appendix:b}
\textbf{Academic Impact.}
DES significantly enhances the reasoning capabilities of small-scale models (\eg, 7B model) through an effective test-time search strategy that leverages additional computation during inference. In contrast, achieving similar improvements via pre-training typically requires scaling up model size and incurring substantial training costs. From this perspective, DES offers a more efficient approach to boosting model performance without retraining or modifying model parameters. This highlights a promising research direction in utilizing computation more strategically—at inference rather than training time—to improve model capabilities, thereby reducing the barriers to deploying strong reasoning models in resource-constrained settings. It also sheds light on the potential of unlocking existing model capacity through smarter inference-time strategies, without the need for costly model retraining. Besides, DES dynamically adjusts the expert configuration during inference, enhancing performance without modifying model parameters. This design offers compelling insights into the synergy between Mixture-of-Experts (MoE) architectures and test-time scaling. By demonstrating how MoE models can be effectively scaled post-training, DES introduces a practical pathway for deploying models more efficiently. We believe this approach will inspire future research in controllable reasoning and adaptive inference, further unlocking the potential of MoE architectures in academic applications.

\textbf{Social Impact.}
The ultimate objective of DES is to substantially enhance the reasoning capabilities of language models through test-time scaling, rather than relying on scaling up model size. By enabling small models (\eg, 7B model) to perform competitively on complex reasoning tasks, DES offers a pathway to making powerful AI models more accessible and deployable in memory-constrained environments(\eg, personal devices). This shift reduces the dependency on large-scale cloud infrastructures and enables real-time and offline inference without internet connectivity, effectively eliminating latency caused by network transmission. This paradigm holds the potential to democratize access to advanced AI capabilities, especially in regions or scenarios where computational resources and reliable network connections are limited. It empowers individuals to benefit from intelligent assistants locally and securely, fostering greater privacy and responsiveness in AI applications.

\section{Implementation Details}
\label{appendix:c}

\subsection{Prompts}
To enable the model to solve reasoning problems, we use the following prompts. We record the prompt as chat format
\begin{tcolorbox}[myhighlight]
    [
    
    \hspace{4pt} \{"role":"system", "content": \textcolor{blue}{[system\_prompt]}\},
    
    \hspace{4pt} \{"role":"user", "content": \textcolor{blue}{[question]}\},
    
    \hspace{4pt} \{"role":"assistant", "content":\}
    
    ]
\end{tcolorbox}

For mathematical task, \textcolor{blue}{[system\_prompt]} is 
\begin{tcolorbox}[myhighlight]
Solve the following math problem efficiently and clearly: \\

- For simple problems (2 steps or fewer):\\
Provide a concise solution with minimal explanation.\\
- For complex problems (3 steps or more):\\
Use this step-by-step format:\\
\newline
\#\# Step 1: [Concise description]\\

[Brief explanation and calculations] \\

\#\# Step 2: [Concise description]\\

[Brief explanation and calculations] \\

...\\

Regardless of the approach, always conclude with:\\

Therefore, the final answer is: \$\textbackslash \textbackslash boxed\{answer\}\$. I hope it is correct.\\

Where [answer] is just the final number or expression that solves the problem.
\end{tcolorbox}

For code-generation tasks, \textcolor{blue}{[system\_prompt]} is

\begin{tcolorbox}[myhighlight]
You are a code assistant.

\#\#\# Instruction: You will be given a question (problem specification) and will generate a correct Python program that matches the specification.
\end{tcolorbox}

For knowledge-domain tasks, \textcolor{blue}{[system\_prompt]} is

\begin{tcolorbox}[myhighlight]
Solve the following question step by step.
\end{tcolorbox}

\subsection{Inference Details}
\label{inference_detail}

\textbf{Deployment.} In our experiments, we set the candidates counts retained per step $M=\frac{N}{4}$ where $N$ denotes the predefined total number of final candidate solutions, maximum steps $T=10$ for all benchmarks. The initial candidate numbers of activated experts is $[4, 5, 6, 7, 8, 9, 10, 11]$ (\textit{s.t.} $n=8$), where all policy models have a default of 8 activated experts. The temperature is set to 0.8 for all search strategies. All experiments were conducted on 4 NVIDIA RTX A6000 GPUs.

\textbf{Inference Framework.}
All policy models are served using the \texttt{vLLM}~\cite{kwon2023efficient} inference framework to ensure efficient and scalable generation and all reward models are served using the \texttt{tranformers}~\cite{wolf-etal-2020-transformers} framework.

\textbf{Answer extraction for mathematical tasks.}
To extract the model's predicted answer from the raw output string, we implement a post-processing function that handles various formats and common answer patterns. The function locates the final answer using segments following ``\texttt{final answer is \$}'' or enclosed within ``\texttt{boxed\{\}}''.

\textbf{Answer extraction for mathematical tasks.}  
To extract the model's predicted answer from its raw output in mathematical tasks, we implement a post-processing function that accounts for diverse output formats and common answer patterns. The function identifies the final answer either from segments following ``\texttt{final answer is \$}'' or from content enclosed within ``\texttt{boxed\{\}}'', ensuring consistent and accurate extraction for evaluation.

\textbf{Answer extraction for code-generation tasks.}
In code-generation tasks, models typically produce their outputs within markdown-style code blocks, often prefixed with a language identifier such as \verb|```python| and ended with \verb|```|. To obtain the executable answer, we extract the content inside these code blocks. Specifically, we identify the opening and closing code block delimiters (\verb|```|) and retrieve all text in between, ignoring any surrounding explanations or comments. This ensures that only the generated code is used as the final answer, which can then be executed or evaluated.

\textbf{Answer extraction for knowledge-domain tasks.}
For tasks in the knowledge domain, the model is prompted to produce its final answer enclosed within a \texttt{<solution>} tag, i.e., \texttt{<solution> ... </solution>}. During evaluation, we extract the content between these tags using a regular expression pattern such as \texttt{<solution>(.*?)</solution>} and treat it as the model's predicted answer. This approach ensures consistent and precise extraction of the answer from the model's textual output.

\section{License}
\label{appendix:d}

The code will be publicly accessible upon acceptance. We use standard licenses from the community. We include the following licenses for the codes, datasets and models we used in this paper.

\begin{enumerate}
  \item \textbf{Benchmarks} \\
  SVAMP: \href{https://github.com/arkilpatel/SVAMP}{MIT} \\
  GSM8K: \href{https://github.com/openai/grade-school-math/blob/master/LICENSE}{MIT} \\
  MATH500: \href{https://github.com/hendrycks/math/blob/master/LICENSE}{MIT} \\
  AIME: \href{https://huggingface.co/datasets/choosealicense/licenses/blob/main/markdown/apache-2.0.md}{Apache}\\
  HumanEval: \href{https://github.com/openai/human-eval/blob/master/LICENSE}{MIT}\\
  LiveCodeBench: \href{https://github.com/LiveCodeBench/LiveCodeBench/blob/main/LICENSE}{MIT}\\
  LiveBench: \href{https://github.com/LiveBench/LiveBench/blob/main/LICENSE}{Apache}\\

  \item \textbf{Models} \\
  Qwen3-30B-A3B: \href{https://huggingface.co/Qwen/Qwen3-30B-A3B/blob/main/LICENSE}{Apache}\\
  Ling-lite-1.5: \href{https://huggingface.co/datasets/choosealicense/licenses/blob/main/markdown/mit.md}{MIT}\\
  OLMoE-1B-7B-Instruct: \href{https://huggingface.co/datasets/choosealicense/licenses/blob/main/markdown/apache-2.0.md}{Apache} \\
  DeepSeek-V2-Lite-Chat: \href{https://github.com/deepseek-ai/DeepSeek-V2/blob/main/LICENSE-MODEL}{Deepseek} \\
  Llama3.1-8B-PRM-Deepseek-Data: \href{https://github.com/meta-llama/llama/blob/main/LICENSE}{Llama} \\
  Qwen2.5-Math-PRM-7B: \href{https://huggingface.co/Qwen/Qwen2.5-Math-PRM-7B/blob/main/LICENSE}{Qwen}
\end{enumerate}

\section{The Use of Large Language Models (LLMs)}
\label{appendix:e}
During the preparation of this manuscript, Large Language Models were used as a general-purpose writing assistant tool. Specifically, LLMs were employed to polish the language and refine the clarity of the text. The authors take full responsibility for the content of the paper.

\end{document}